# SPEAKER IDENTIFICATION IN EACH OF THE NEUTRAL AND SHOUTED TALKING ENVIRONMENTS BASED ON GENDER-DEPENDENT APPROACH USING SPHMMS

Ismail Shahin

Electrical and Computer Engineering Department

University of Sharjah

P. O. Box 27272

Sharjah, United Arab Emirates

E-mail: ismail@sharjah.ac.ae

**Abstract**

It is well known that speaker identification performs extremely well in the neutral talking environments; however, the identification performance is declined sharply in the shouted talking environments. This work aims at proposing, implementing and testing a new approach to enhance the declined performance in the shouted talking environments. The new proposed approach is based on gender-dependent speaker identification using Suprasegmental Hidden Markov Models (SPHMMs) as classifiers. This proposed approach has been tested on two different and separate speech databases: our collected database and the Speech Under Simulated and Actual Stress (SUSAS) database. The results of this work show that gender-dependent speaker identification based on SPHMMs outperforms gender-independent speaker identification based on the same models and gender-dependent speaker identification based on Hidden Markov Models (HMMs) by about 6% and 8%, respectively. The results obtained based on the proposed approach are close to those obtained in subjective evaluation by human judges.

## 1. Introduction

Speaker recognition, in general, can be subdivided into speaker identification and speaker verification (authentication). Speaker identification focuses on determining the identity of a particular speaker. The applications of speaker identification systems can be seen clearly in criminal investigations to decide the suspected persons produced voices recorded at the site of the crime, they also emerge in civil situations or for the media. These situations involve calls to radio stations, local or other government authorities, insurance companies, observing people by their voices and many other applications [1]. Speaker verification focuses on verifying the claimed identity of a speaker. The applications of speaker verification systems appear in: banking transactions using a telephone network, database access services, security control for constrained information areas and remote log on to computers.

Speaker identification systems are divided into two sets: "closed set" and "open set". In the "closed set", it is assumed that the unknown speaker is in the set of known speakers, while in the "open set", the unknown speaker may or may not be in the set of known speakers. Speaker identification systems typically operate in one of two texts: text-dependent (fixed-text) or text-independent (free-text). In text-dependent, the unknown speakers must speak the same prescribed text for both training and testing (identification). On the other hand, in text-independent, the unknown speakers are allowed to read any text during both training and testing.

## 2. Literature Review and Motivation

Some previous studies have focused on speech and speaker recognition in stressful talking environments. The authors of *Ref.* [2] evaluated the effectiveness of traditional features in recognition of speech under stress and formulated new features which are shown to enhance stressed speech recognition [2]. The authors of *Ref.* [3] applied hidden Markov models (HMMs) in the study of nonlinear feature based classification of speech under stress [3]. The author of *Ref.* [4] focused on talker-stress-induced intra-word variability and an algorithm that pays off for the systematic changes observed based on HMMs trained by speech tokens in distinct talking conditions [4].

Although there are some studies focused on speech and speaker recognition in stressful talking environments, there are very few studies focused on speaker recognition in shouted talking environments [5], [6], [7], [8]. Shouted talking environments are described as the talking environments in which when speakers shout, their intention is to generate a very loud acoustic signal, either to raise its extent of transmission or its ratio to background noise [5], [6], [7]. In four of his earlier studies, the author of *Refs.* [5], [6], [7], [8] concentrated on enhancing speaker identification performance in shouted talking environments based on each of the: Second-Order Hidden Markov Models (HMM2s) [5], Second-Order Circular Hidden Markov Models (CHMM2s) [6], Suprasegmental Hidden Markov Models (SPHMMs) [7] and Second-Order Circular Suprasegmental Hidden Markov Models (CSPHMM2s) [8]. In one of his most recent studies, the author of *Ref.* [9] shed the light on improving speaker identification performance in shouted talking environments based on gender-dependent speaker identification approach using HMMs [9].

Speaker identification systems in shouted talking environments can be applied in the applications of talking condition identification systems. These systems can be employed in the applications of medicine where computerized stress categorization and evaluation methods can be utilized by psychiatrists to assist in quantitative objective evaluation of patients who suffer assessment. Such systems can also be exploited in the applications of talking condition intelligent automated systems in call-centers.

The contribution of this work is focused on proposing, implementing and evaluating gender-dependent speaker identification approach in each of the neutral and shouted talking environments based on SPHMMs as classifiers. Two separate and distinct speech databases have been used in this work to evaluate the proposed approach. The first database is the collected speech database and the second one is the Speech Under Simulated and Actual Stress (SUSAS) database. Gender-dependent speaker identification gives more accurate and more specific information about the identity of speakers than gender-independent speaker identification in: criminal investigations to decide the suspected persons uttered the voice during crimes, calls to radio stations and monitoring people by their voices. In call centers of some conservative societies, automatic dialog systems with the capability of recognizing genders are prefered over those without this capability.

The organization of the paper is as follows. The next section overviews SPHMMs. Section 4 describes the two speech databases used in this work. Section 5 is committed to discussing the proposed approach and the experiments. Section 6 discusses the results that are obtained in this work. Concluding remarks are given in Section 7.

## 3. Overview of Suprasegmental Hidden Markov Models

SPHMMs have been developed, implemented and evaluated by the author of *Ref.* [7] for speaker identification systems in shouted talking environments [7]. SPHMMs have proven to be superior models over HMMs for speaker identification in such talking environments [7]. SPHMMs have the ability to sum up several states of HMMs into what is named suprasegmental state. Suprasegmental state possesses the ability to look at the observation sequence through a larger window. This suprasegmental state permits observations at appropriate rates for the case of modelling. Prosodic information, for example, can not be found at a rate that is used for acoustic modelling. The main acoustic parameters that express prosody are: fundamental frequency, intensity and duration of speech segments [10]. The prosodic features of a unit of speech are characterized as suprasegmental features because they have impact on all the segments of the unit of speech. Therefore, prosodic events at the levels of: phone, syllable, word and utterance are represented using suprasegmental states; on the other hand, acoustic events are represented using conventional hidden Markov states.

Within HMMs, prosodic and acoustic information can be combined and integrated as given by the following formula [11]:

$$log\ \mathrm{P}\left(\lambda^v, \Psi^v | \mathrm{O}\right) = (1-\alpha).\ log\ \mathrm{P}\left(\lambda^v | \mathrm{O}\right) + \alpha.\ log\ \mathrm{P}\left(\Psi^v | \mathrm{O}\right) \qquad (1)$$

where $\alpha$ is a weighting factor. When [8]:

$$\begin{cases} 0.5 > \alpha > 0 & \text{biased towards acoustic model} \\ 1 > \alpha > 0.5 & \text{biased towards prosodic model} \\ \alpha = 0 & \text{biased completely towards acoustic model and} \\ & \text{no effect of prosodic model} \\ \alpha = 0.5 & \text{no biasing towards any model} \\ \alpha = 1 & \text{biased completely towards prosodic model and} \\ & \text{no impact of acoustic model} \end{cases} \quad (2)$$

$\lambda^v$: is the $v^{th}$ acoustic model.

$\Psi^v$: is the $v^{th}$ suprasegmental model.

$O$: is the observation vector or sequence of an utterance.

$P(\lambda^v | O)$ and $P(\Psi^v | O)$ can be calculated using Bayes theorem as given in *Eqs.* (3) and (4), respectively [12]:

$$P(\lambda^v | O) = \frac{P(O | \lambda^v) P_0(\lambda^v)}{P(O)} \quad (3)$$

$$P(\Psi^v | O) = \frac{P(O | \Psi^v) P_0(\Psi^v)}{P(O)} \quad (4)$$

where $P_0(\lambda^v)$ and $P_0(\Psi^v)$ are the priori distribution of the acoustic model and the suprasegmental model, respectively. The parameter priori distribution describes the statistics of the parameters of relevance before any measurement is made. More information about SPHMMs can be obtained from the *Refs.* [7] and [8].

## 4. Speech Databases

In this work, two different speech databases have been separately used to test the proposed approach. The first database is the collected database and the second one is the SUSAS database. The two databases have been used as "closed set" in the present work.

### 4.1 The Collected Speech Database

In this database, eight sentences were produced under each of the neutral and shouted talking environments. These sentences are:

1) *He works five days a week.*
2) *The sun is shining.*
3) *The weather is fair.*
4) *The students study hard.*
5) *Assistant professors are looking for promotion.*
6) *University of Sharjah.*
7) *Electrical and Computer Engineering Department.*
8) *He has two sons and two daughters.*

Twenty five male and twenty five female healthy adult native speakers of American English were separately asked to generate the eight sentences. The fifty speakers were untrained (uttering sentences naturally) to avoid overstated expressions. The speakers were asked to portray each sentence five times in one session (training session) and four times in another separate session (evaluation session) in each of the neutral and shouted talking conditions.

The collected database was captured by a speech acquisition board using a 12-bit linear coding A/D converter and sampled at a sampling rate of 12 kHz. This database was a 12-bit per sample linear data.

**4.2 SUSAS Database**

SUSAS database was designed originally for speech recognition under neutral and stressful talking conditions [13], [14]. In this work, isolated words recorded at 8 kHz sampling rate under each of the neutral and angry talking conditions have been used in this database [14]. Angry talking condition has been used as a substitute to the shouted talking condition because the shouted talking condition can not be completely separated from the angry talking condition in real life [5]. Thirty distinct utterances uttered by seven speakers (four male and three female) in each of the neutral and angry talking conditions have been chosen to evaluate the proposed approach.

In this work, Mel Frequency Cepstral Coefficients (MFCCs) have been used as features to represent the phonetic content of speech signals in each database. MFCCs have been most commonly used in each of stressful speech recognition and stressful speaker recognition fields because of their performance superiority over other features in the two fields and because of providing a high-level approximation of human auditory perception [8], [15], [16], [17], [18], [19], [20]. These spectral features have also been found to be useful in the classification of stress in speech [21].

MFCCs were computed with the help of a psycho acoustically motivated filter bank, followed by logarithmic compression and Discrete Cosine Transform (DCT). These coefficients can be computed as given in the following formula [22]:

$$C(n) = \sum_{m=1}^{M} \left\{ [log \ Y(m)] \ cos\left[ \frac{\pi n}{M} \left(m - \frac{1}{2}\right) \right] \right\} \qquad (5)$$

where *Y(m)* are the outputs of an *M*-channel filter bank. Fig. 1 illustrates a block diagram of generating MFCCs. In this work, a 16-dimension MFCC feature analysis was used to form the observation vectors in SPHMMs.

Most of the studies performed in the last three decades in the fields of speech recognition and speaker recognition on HMMs have been done using Left-to-Right Hidden Markov Models (LTRHMMs) because phonemes follow strictly the left-to-right sequence [23], [24], [25]. In this work, Left-to-Right Suprasegmental Hidden Markov Models (LTRSPHHMs) have been derived from LTRHMMs. Fig. 2 shows an example of a basic structure of LTRSPHMMs that has been derived from LTRHMMs. In this figure, $q_1$, $q_2$, ..., $q_6$ are conventional hidden Markov states, $p_1$ is a suprasegmental state that consists of $q_1$, $q_2$ and $q_3$, $p_2$ is a suprasegmental state that is made up of $q_4$, $q_5$ and $q_6$, $p_3$ is a suprasegmental state that is composed of $p_1$ and $p_2$, $a_{ij}$ is the transition probability between the *i*th conventional state and the *j*th conventional state and $b_{ij}$ is the transition probability between the *i*th suprasegmental state and the *j*th suprasegmental state.

In this work, the number of conventional states of LTRHMMs, *N*, is nine. The number of mixture components, *M*, is ten per state, with a continuous mixture observation density has been selected for these models. In LTRSPHMMs, the number of suprasegmental states is three. Therefore, each three conventional states of LTRHMMs in the current work are summarized into one suprasegmental state. A continuous mixture observation density has been selected for LTRSPHMMs.

## 5. Gender-Dependent Speaker Identification Approach and the Experiments

Given *n* speakers per gender, the overall proposed approach is illustrated in Fig. 3. This figure shows that gender-dependent speaker identification approach consists of a two-stage recognizer that combines and integrates both gender recognizer and speaker recognizer into one system. The two stages are:

**5.1 Gender Identification Stage**

The first stage of the overall proposed architecture is to identify the gender of the unknown speaker in order to make the output of this stage gender-dependent. The problem of differences in features for the two genders is well-known in the field of speaker recognition [16]. Automatic gender identification, in general, yields high performance with little effort since the output of this recognizer is that the speaker is either a male (M) or a female (F). Therefore, gender identification is a binary classification problem which is generally not complicated.

In one of their studies to enhance speech recognition performance through gender separation, the authors of *Ref.* [26] separated the datasets based on the gender to build gender-dependent HMM for each word [26]. Based on their method, word recognition performance has been significantly improved over the gender-independent method. The authors of *Ref.* [16] preceded their emotion recognizer by a gender recognizer to enhance emotion recognition performance [16]. The author of *Ref.* [9] proposed a new approach called gender-dependent speaker identification based on HMMs to improve the degraded speaker identification performance in shouted talking environments [9].

In this stage, two probabilities per utterance are computed based on SPHMMs and the maximum probability is chosen as the identified gender as given in the following formula:

$$G^* = \arg\max_{2 \geq g \geq 1} \left\{ P\left(O \middle| \Gamma^g\right) \right\} \quad (6)$$

where,

$G^*$: is the index of the identified gender (either *M* or *F*).

$\Gamma^g$: is the $g^{th}$ SPHMM gender model.

$P\left(O \middle| \Gamma^g\right)$: is the probability of the observation sequence *O* that belongs to the unknown gender given the $g^{th}$ SPHMM gender model.

In the training session of this stage and using the collected database, male gender model has been derived using the twenty five male speakers uttering the same sentence in each of the neutral and shouted talking environments, while female gender model has been constructed using the twenty five female speakers generating the same sentence in each of the neutral and shouted talking environments. In this session, the total number of utterances that has been used in the collected database and SUSAS database is 4000 and 420, respectively. The training session of SPHMM gender model is very similar to the training session of conventional HMM gender model. In the training session of SPHMM gender model, suprasegmental gender models are trained on top of acoustic gender models of HMMs.

**5.2 Speaker Identification Stage**

The second stage of the proposed approach is to identify the unknown speaker given that his/her gender was identified. This stage is gender-specific speaker identification. In this stage, $n$ probabilities per gender are computed based on SPHMMs and the maximum probability is selected as the identified speaker as given in the following formula:

$$S^* = \arg \max_{n \geq s \geq 1} \left\{ P\left(O \mid G^*, \Theta^s\right) \right\} \quad (7)$$

where,

$S^*$: is the index of the identified speaker.

$P\left(O \mid G^*, \Theta^s\right)$: is the probability of the observation sequence $O$ that belongs to the unknown speaker given the $s^{th}$ SPHMM speaker model $\left(\Theta^s\right)$ and the identified gender.

The $s^{th}$ SPHMM speaker model has been derived using five of the nine utterances per speaker per sentence in the neutral talking environments using the collected database. In this session, the total number of utterances that has been used in the collected database and the SUSAS database is 2000 and 210, respectively. The training session of SPHMM speaker model is very similar to the training session of conventional HMM speaker model. In the training session of SPHMM speaker model, suprasegmental speaker model is trained on top of acoustic HMM speaker model.

In the evaluation (identification) session, each one of the twenty five male speakers and the twenty five female speakers used four utterances per the same sentence (text-dependent) in each of the neutral and shouted talking environments. The total number of utterances that has been used in this session using the collected database and SUSAS database is 3200 and 420, respectively. A block diagram of this stage is shown in Fig. 4.

## 6. Results and Discussion

In this work, the proposed approach has been tested using separately each of the collected and SUSAS databases when the weighting factor ($\alpha$) is equal to 0.5 to avoid biasing towards any model.

Using the collected speech database, automatic gender identification performance based on SPHMMs (output of the gender identification recognizer) is 98.3% and 93.2% in the neutral and shouted talking environments, respectively. Using SUSAS database, gender identification performance based on the same models is 99.5% and 94.3% in the neutral and angry talking environments, respectively.

Gender identification performance obtained in this work is higher than that reported in some previous studies. The authors of *Ref.* [16] obtained gender identification performance of 90.26% and 91.85% using Berlin and SmartKom German databases, respectively [16]. The authors of *Ref.* [27] reported gender identification performance of 92% in neutral talking environments [27]. On the other hand, the authors of *Ref.* [26] achieved better results than the results obtained in this work in neutral talking environments. They achieved gender identification performance of 100% based on the average pitch method [26]. The author of *Ref.* [9] achieved in one of his studies, based on HMMs, an automatic gender identification performance of 97% and 90.5% in the neutral and shouted talking environments, respectively [9]. Based on the same models and using SUSAS database, he obtained automatic gender identification performance of 98% and 92% in the neutral and angry talking environments, respectively [9]. Fig. 5 illustrates relative improvement of using SPHMMs over HMMs in gender identification recognizer using each of the collected and SUSAS databases. It is apparent from this figure that gender identification performance in each of the neutral and shouted/angry talking environments based on SPHMMs has been insignificantly improved compared to that based on HMMs.

Using the collected database, speaker identification performance of the overall system based on the proposed approach and using SPHMMs is 97.8% and 79.2% in the neutral and shouted talking environments, respectively. Using SUSAS database, speaker identification performance of the overall system based on the same approach and using the same models is 98.5% and 80% in the neutral and angry talking environments, respectively.

The author of *Ref.* [9] obtained in one of his work using HMMs 96.5% and 73.5% as speaker identification performance in the neutral and shouted talking environments, respectively, of the overall system based on gender-dependent speaker identification approach [9]. Based on the same approach and models and using SUSAS database, he attained speaker identification performance of 97% and 74% in the neutral and angry talking environments, respectively [9]. Fig. 6 demonstrates the relative improvement of using SPHMMs over HMMs in speaker identification performance based on the proposed architecture using each of the collected and SUSAS databases. It is evident from this figure that SPHMMs significantly enhance speaker identification performance in the shouted/angry talking environments based on the proposed approach compared to that using HMMs based on the same approach.

Combining both gender and speaker information (gender-dependent speaker identificaion system) based on the proposed approach leads to a higher speaker identification performance in each of the neutral and shouted/angry talking environments than that achieved in some previous studies:

1) Gender-independent speaker identification performance based on SPHMMs. The author of *Ref.* [7] achieved in one of his studies in the neutral talking environments a 99% for each of male and female speaker identification performances. On the other hand, male and female speaker identification performance in the shouted talking environments is 74% and 76%, respectively [7]. In the shouted talking environments, it is apparent that gender-dependent speaker identification approach based on SPHMMs is superior to gender-independent speaker identification approach based on the same models by 6%.

2) Gender-independent speaker identification performance based on HMM2s. In one of his previous studies, the author of *Ref.* [5] got 92% and 96% as a male and female speaker identification performance in the neutral talking environments, respectively. In the shouted talking environments, male and female speaker identification performance is 57% and 61%, respectively [5]. It is evident that gender-dependent speaker identification approach based on SPHMMs outperforms gender-independent speaker identification approach based on HMM2s by 34.7% in the shouted talking environments.

3) Gender-independent speaker identification performance based on CHMM2s. The author of *Ref.* [6] reported in one of his studies male and female speaker identification performance in the neutral talking environments a 94% and 97%, respectively. In the shouted talking environments, male and female speaker identification performance is 71% and 73%, respectively [6]. It is clear that gender-dependent speaker identification architecture based on SPHMMs leads gender-independent speaker identification architecture based on CHMM2s by 10.4% in the shouted talking environments.

The current proposed approach has been evaluated for different values of the weighting factor ($\alpha$). Fig. 7 and Fig. 8 show speaker identification performance in each of the neutral and shouted/angry talking environments based on the proposed approach for different values of $\alpha$ (0.0, 0.1, 0.2, …, 0.9, 1.0) using the collected database and SUSAS database, respectively. The two figures indicate that increasing the value of the weighting factor has a significant impact on improving speaker identification performance in the shouted/angry talking environments. The two figures illustrate clearly that in the neutral talking environments the identification performance has been insignificantly enhanced with the increase of the weighting factor. Therefore, it is evident, based on the proposed approach, that suprasegmental hidden Markov models have more influence than acoustic hidden Markov models on speaker identification performance in shouted talking environments.

An informal subjective evaluation based on the proposed architecture using the collected speech database was carried out with ten nonprofessional listeners (human judges). A total of 400 utterances (twenty five speakers per gender, two talking environments and four sentences only) were used in this evaluation. During the assessment, the listeners were asked to answer two questions for every test utterance. The two questions were: identify the unknown gender and identify the unknown speaker. The results of this evaluation were encouraging. Gender identification performance was 95.1% and 92.2% in the neutral and shouted talking environments, respectively. In the neutral and shouted talking environments, speaker identification performance was 94.6% and 77.1%, respectively. These human results were close to the achieved results based on the proposed approach using SPHMMs.

## 7. Concluding Remarks

Some conclusions can be drawn from this work. Firstly, combining and integrating gender recognizer and speaker recognizer into one system based on SPHMMs lead to a significant enhancement in speaker identification performance over that of gender-independent speaker identification in the shouted/angry talking environments based on each of SPHMMs, HMM2s and CHMM2s. Secondly, SPHMMs outperform HMMs for gender-dependent speaker identification approach. Finally, speaker identification performance using speaker's gender based on the proposed approach is limited in the shouted/angry talking environments. Speaker identification performance based on the proposed approach is the resultant of two performances. The reasons of the limitations are:

a) Gender identification recognizer does not give perfect results. Gender identification performance is less than 100%.

b) The unknown speaker in the speaker identification recognizer is not 100% correctly identified.

There are some limitations when the SUSAS database has been used to test the proposed approach. First, the number of speakers that has been used in this database is limited to seven. Second, angry talking condition has been used as an alternative to the shouted talking condition. Third, isolated words have been used in this database instead of sentences (in the collected database) to assess the proposed approach.

**References**

[1] S. Furui, Speaker-dependent-feature-extraction, recognition, and processing techniques, *Speech Communication*, *10*, March 1991, 505-520.

[2] S. E. Bou-Ghazale & J. H. L. Hansen, A comparative study of traditional and newly proposed features for recognition of speech under stress, *IEEE Transaction on Speech and Audio Processing*, *8* (4), July 2000, 429-442.

[3] G. Zhou, J. H. L. Hansen, & J. F. Kaiser, Nonlinear feature based classification of speech under stress, *IEEE Transaction on Speech and Audio Processing*, *9* (3), March 2001, 201-216.

[4] Y. Chen, Cepstral domain talker stress compensation for robust speech recognition, *IEEE Transaction on ASSP*, *36* (4), April 1988, 433-439.

[5] I. Shahin, Improving speaker identification performance under the shouted talking condition using the second-order hidden Markov models, *EURASIP Journal on Applied Signal Processing*, *5* (4), March 2005, 482-486.

[6] I. Shahin, Enhancing speaker identification performance under the shouted talking condition using second-order circular hidden Markov models, *Speech Communication*, *48* (8), August 2006, 1047-1055.

[7] I. Shahin, Speaker identification in the shouted environment using suprasegmental hidden Markov models, *Signal Processing Journal*, *88* (11), November 2008, 2700-2708.

[8] I. Shahin, Employing second-order circular suprasegmental hidden Markov models to enhance speaker identification performance in shouted talking environments, *EURASIP Journal on Audio, Speech, and Music Processing*, *2010*, Article ID 862138, 10 pages, 2010. doi:10.1155/2010/862138.

[9] I. Shahin, Gender-dependent speaker identification under shouted talking condition, *3$^{rd}$ International Conference on Communication, Computer and Power (ICCCP'09)*, Muscat, Oman, February 2009, 332-336.

[10] J. Adell, A. Benafonte, & D. Escudero, Analysis of prosodic features: towards modeling of emotional and pragmatic attributes of speech," *XXI Congreso de la Sociedad Española para el Procesamiento del Lenguaje Natural*. SEPLN, Granada, Spain, September 2005.

[11] T. S. Polzin & A. H. Waibel, Detecting emotions in Speech, *Cooperative Multimodal Communication*, *Second International Conference 1998*, CMC 1998.

[12] L. R. Rabiner & B. H. Juang, *Fundamentals of speech recognition* (Englewood Cliffs, NJ: Prentice Hall, 1993).

[13] J.H.L. Hansen & S. Bou-Ghazale, Getting started with SUSAS: A speech under simulated and actual stress database, *EUROSPEECH-97: Inter. Conf. On Speech Communication and Technology*, Rhodes, Greece, September 1997, 1743-1746.

[14] http://www.ldc.upenn.edu/Catalog/CatalogEntry.jsp?catalogId=LDC99S78

[15] H. Bao, M. Xu, & T. F. Zheng, Emotion attribute projection for speaker recognition on emotional speech, *INTERSPEECH 2007*, Antwerp, Belgium, August 2007, 758-761.

[16] T. Vogt & E. Andre, Improving automatic emotion recognition from speech via gender differentiation, *Proceedings of Language Resources and Evaluation Conference (LREC 2006),* Genoa, Italy, 2006.

[17] J. Dai, Isolated word recognition using Markov chain models, *IEEE Trans. on Speech and Audio Processing Journal*, *3* (6), November 1995, 458-463.

[18] O. W. Kwon, K. Chan, J. Hao, & T. W. Lee, Emotion recognition by speech signals, *8th European Conference on Speech Communication and Technology 2003*, Geneva, Switzerland, September 2003, 125-128.

[19] I. Luengo, E. Navas, I. Hernaez, & J. Sanches, Automatic emotion recognition using prosodic parameters, *INTERSPEECH 2005*, Lisbon, Portugal, September 2005, 493-496.

[20] T. H. Falk & W. Y. Chan, Modulation spectral features for robust far-field speaker identification, *IEEE Transactions on Audio, Speech and Language Processing*, *18* (1), January 2010, 90-100.

[21] J. H. L. Hansen & B. D. Womack, Feature analysis and neural network-based classification of speech under stress, *IEEE Trans. on Speech and Audio Processing Journal*, *4* (4), July 1996 , 307-313.

[22] T. Kinnunen & H. Li, An overview of text-independent speaker recognition: from features to supervectors, *Speech Communication*, *52* (1), January 2010, 12-40.

[23] T. L. Nwe, S. W. Foo, & L. C. De Silva, Speech emotion recognition using hidden Markov models, *Speech Communication*, *41* (4), November 2003, 603-623.

[24] D. Ververidis & C. Kotropoulos, Emotional speech recognition: resources, features, and methods, Speech *Communication*, *48* (9), September 2006, 1162-1181.

[25] L. T. Bosch, Emotions, speech and the ASR framework, *Speech Communication*, *40* (1-2), April 2003, 213-225.

[26] W. H. Abdulla & N. K. Kasabov, Improving speech recognition performance through gender separation, *Artificial Neural Networks and Expert Systems International Conference (ANNES)*, Dunedin, New Zealand, 2001, 218–222.

[27] H. Harb & L. Chen, Gender identification using a general audio classifier, *International Conference on Multimedia and Expo 2003 (ICME '03)*, July 2003, II – (733 – 736).

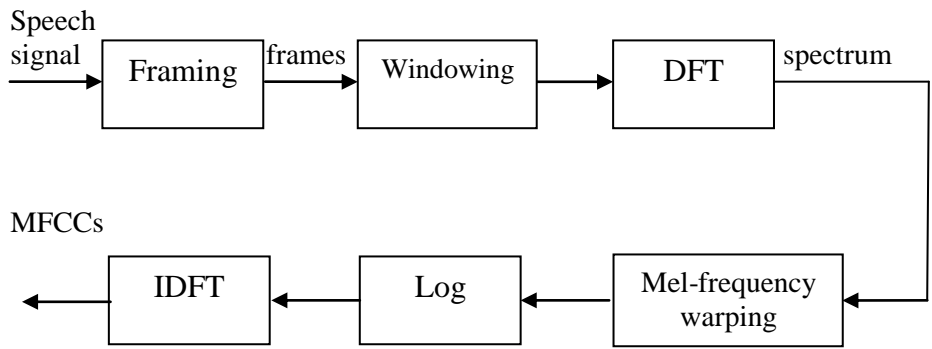

Fig. 1. Generating MFCCs

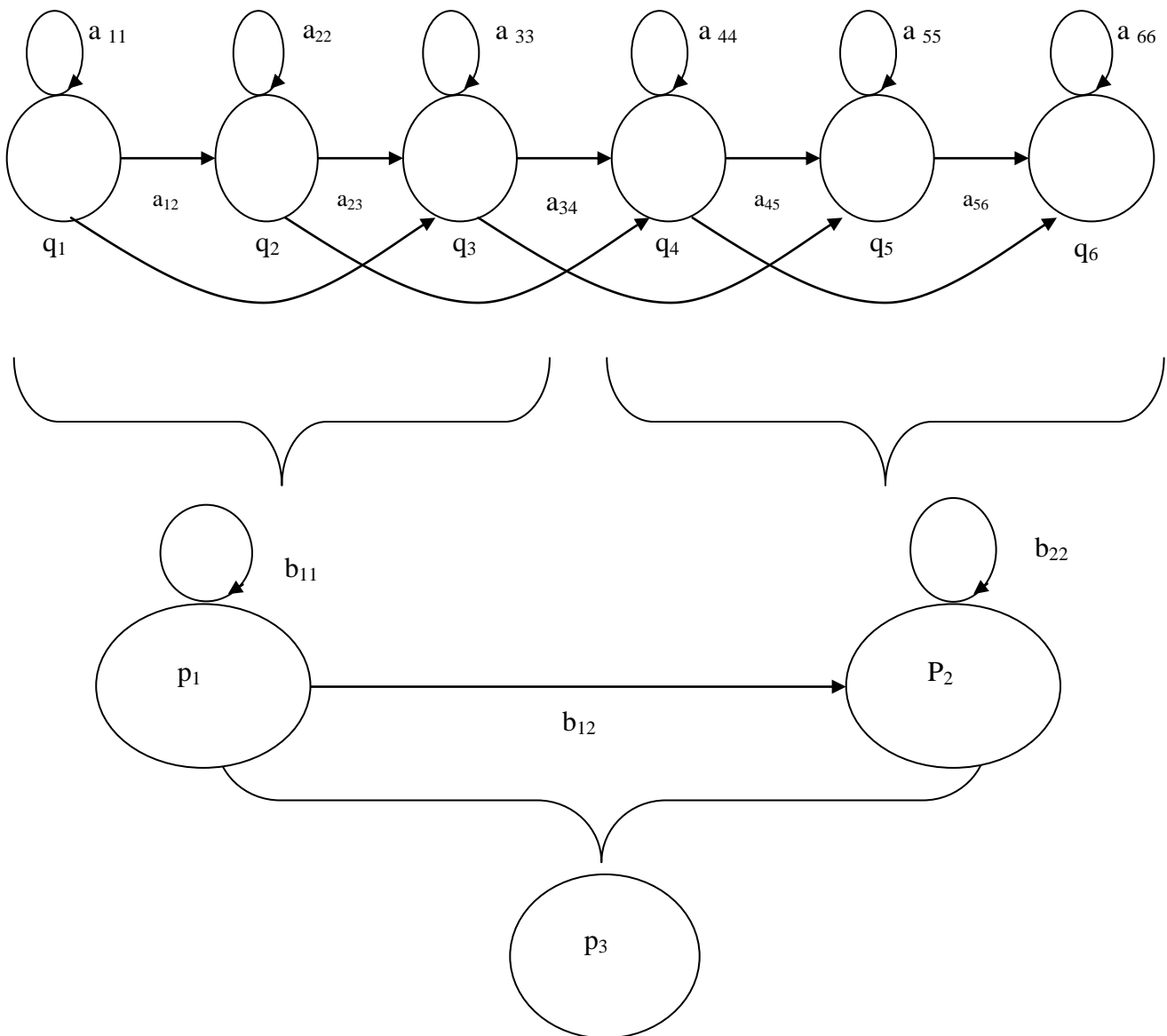

Fig. 2. Basic structure of LTRSPHMMs

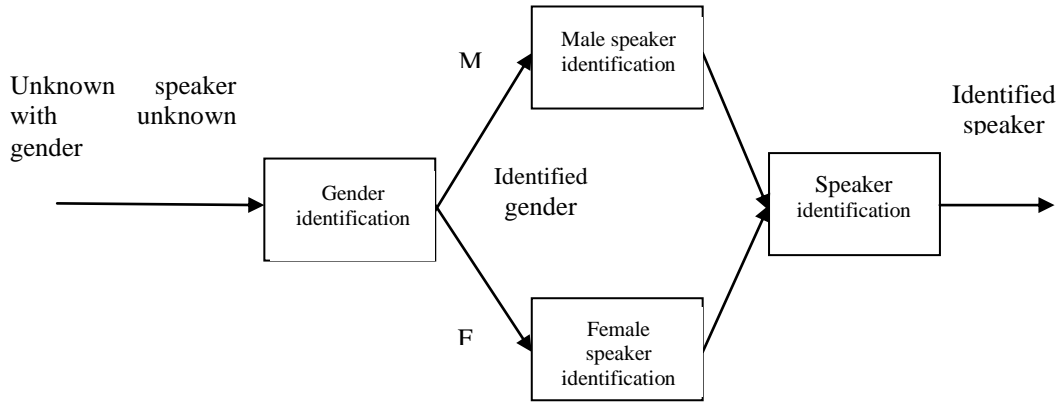

Fig. 3. Block diagram of the overall proposed approach

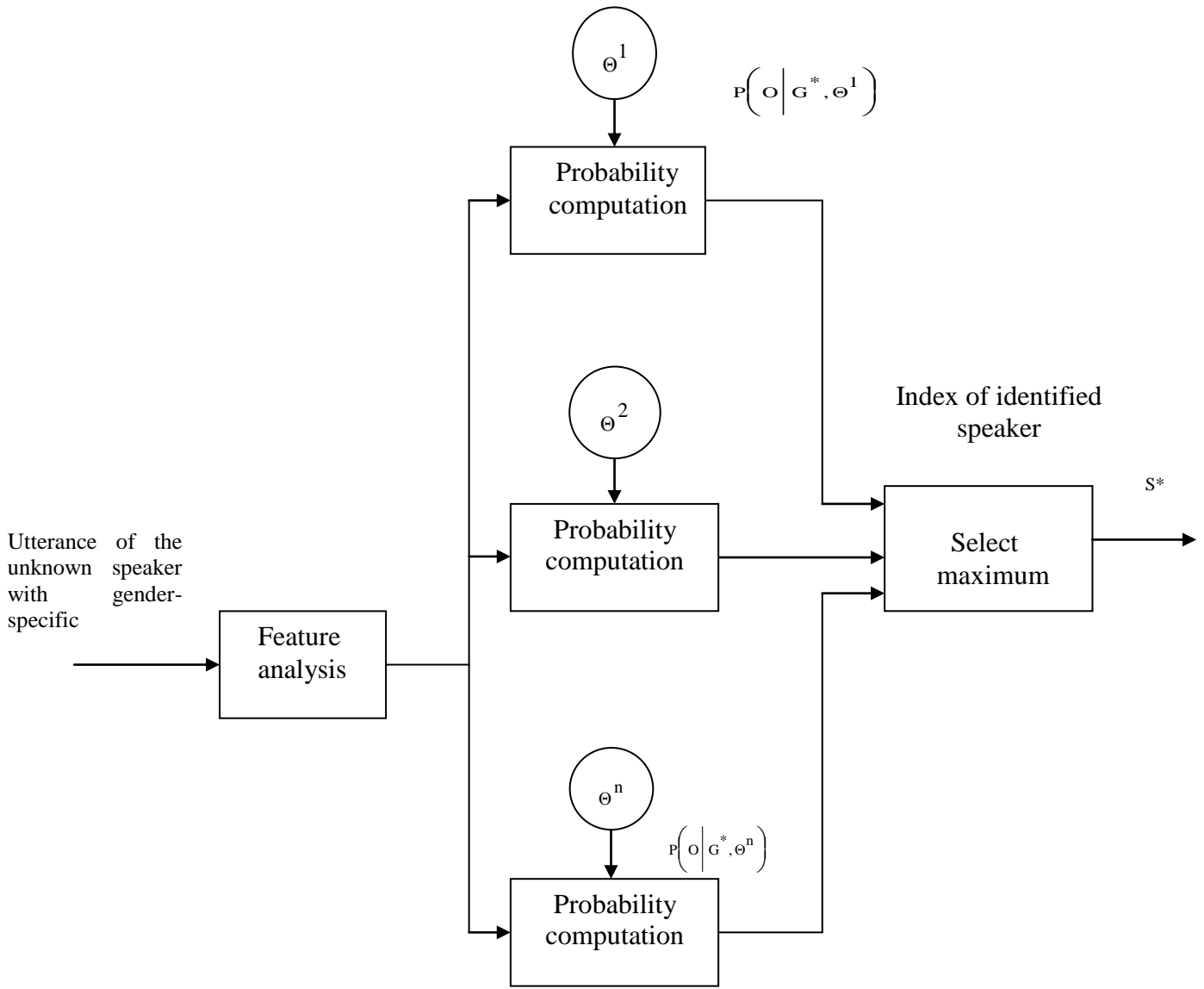

Fig. 4. Block diagram of speaker identification stage

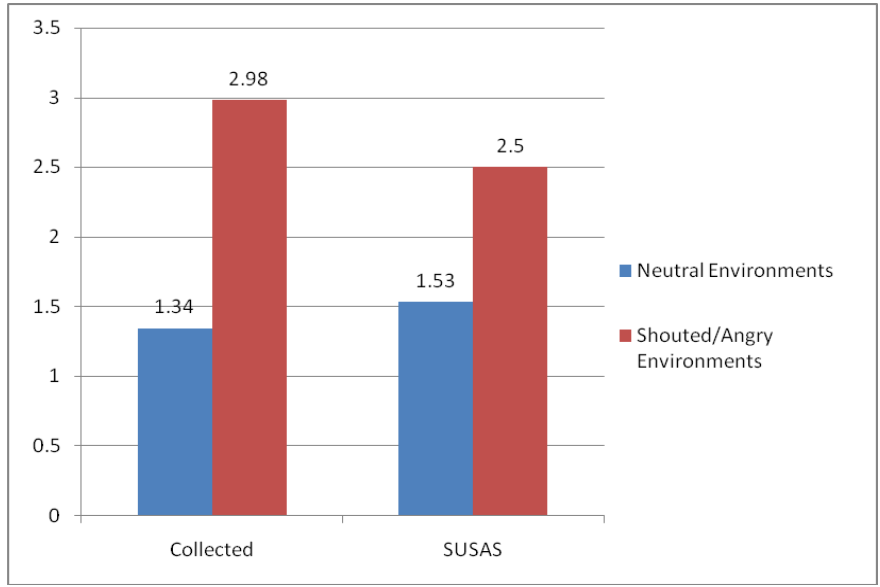

Fig. 5. Relative improvement (%) of using SPHMMs over HMMs in gender identification recognizer using each of the collected and SUSAS databases

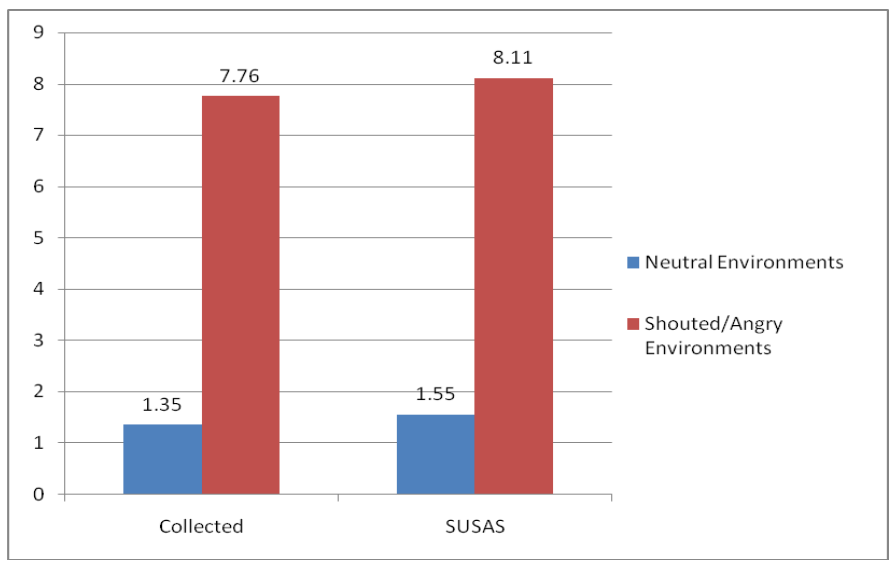

Fig. 6. Relative improvement (%) of speaker identification performance using SPHMMs over HMMs based on the proposed approach using each of the collected and SUSAS databases

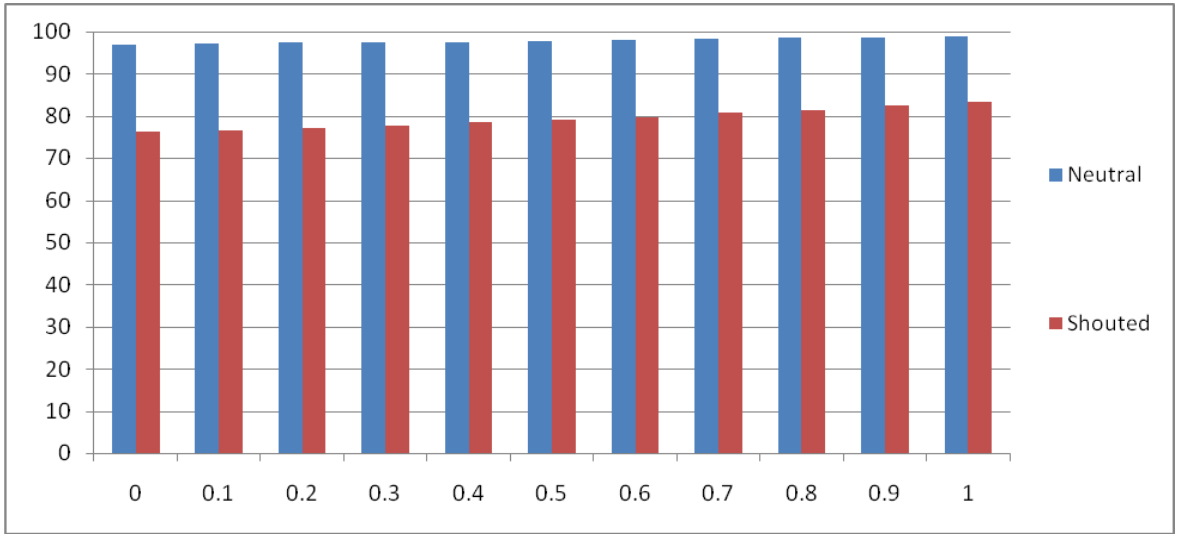

Fig. 7. Speaker identification performance (%) versus the weighting factor $\alpha$ based on the proposed approach using the collected database

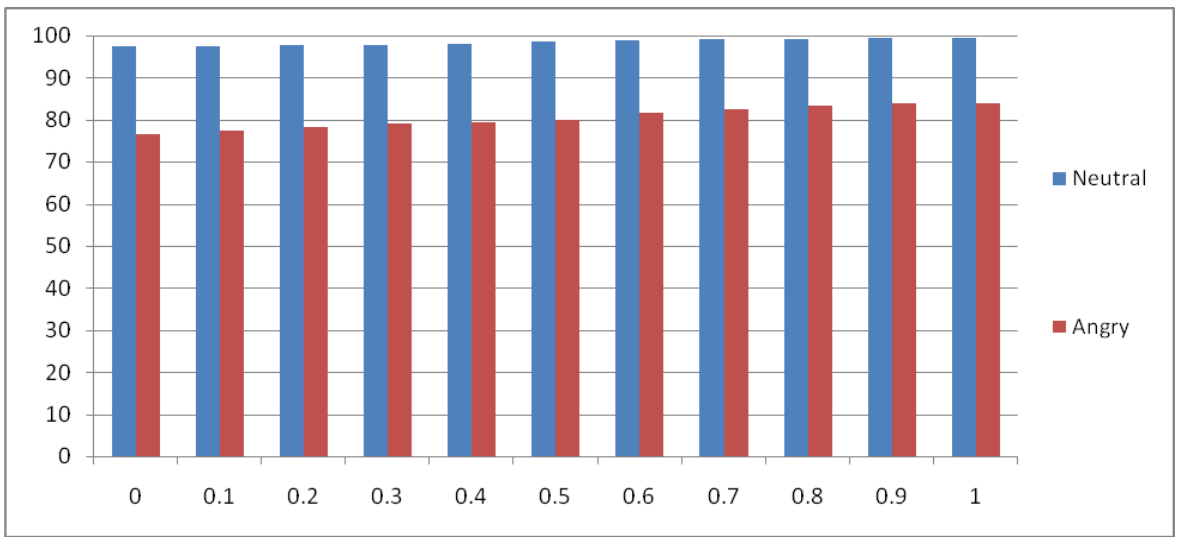

Fig. 8. Speaker identification performance (%) versus the weighting factor $\alpha$ based on the proposed approach using SUSAS database